\documentclass[sigconf]{acmart}

\usepackage{booktabs} 
\usepackage{multirow}
\usepackage{svg}
\usepackage{graphicx}
\usepackage[table,xcdraw]{xcolor}
\usepackage{subcaption}

\setcopyright{rightsretained}

\renewcommand\footnotetextcopyrightpermission[1]{}

\acmDOI{}

\acmISBN{}

\acmConference[PCSC2026]{Philippine Computing Science Congress}{April 2026}{Davao, Philippines}
\acmYear{2026}
\copyrightyear{2026}

\acmArticle{}
\acmPrice{}

\editor{}
\editor{}
\editor{}

\begin{document}
\title{Mixture of Experts with Soft Nearest Neighbor Loss:\\ Resolving Expert Collapse via Representation Disentanglement}

\author{Abien Fred Agarap}
\email{abien.agarap@dlsu.edu.ph}
\affiliation{
	\institution{De~La~Salle~University}
	\city{Manila}
	\country{Philippines}
}
\author{Arnulfo Azcarraga}
\email{arnulfo.azcarraga@dlsu.edu.ph}
\affiliation{
	\institution{De~La~Salle~University}
	\city{Manila}
	\country{Philippines}
}

\renewcommand{\shortauthors}{Agarap and Azcarraga}

\begin{abstract}
The Mixture-of-Experts (MoE) model uses a set of expert networks that specialize on subsets of a dataset under the supervision of a gating network. A common issue in MoE architectures is ``expert collapse'' where overlapping class boundaries in the raw input feature space cause multiple experts to learn redundant representations, thus forcing the gating network into rigid routing to compensate. We propose an enhanced MoE architecture that utilizes a feature extractor network optimized using Soft Nearest Neighbor Loss (SNNL) prior to feeding input features to the gating and expert networks. By pre-conditioning the latent space to minimize distances among class-similar data points, we resolve structural expert collapse which results to experts learning highly orthogonal weights. We employ Expert Specialization Entropy and Pairwise Embedding Similarity to quantify this dynamic. We evaluate our experimental approach across four benchmark image classification datasets (MNIST, FashionMNIST, CIFAR10, and CIFAR100), and we show our SNNL-augmented MoE models demonstrate structurally diverse experts which allow the gating network to adopt a more flexible routing strategy. This paradigm significantly improves classification accuracy on the FashionMNIST, CIFAR10, and CIFAR100 datasets.
\end{abstract}



\maketitle

\section{Introduction and Related Works}
An artificial neural network solves a given problem by learning to approximate the function that describes the relationship between dataset features $x$ and labels $y$ in a given training data. This is accomplished by training with retro-propagation of its output errors \cite{rumelhart1985learning} and then adjusting its weights based on such error signals. Neural networks can be used for a plethora of tasks such as image classification \cite{krizhevsky2012imagenet}, language translation \cite{cho2014learning}, and speech recognition \cite{graves2013speech}. It learns to solve these problems by breaking them down to learn the best representations for the input features, and then performing the given task on the resulting learned representations. This representation learning capability is among the advantages of neural networks over other machine learning algorithms. 

\subsection{Class Neighborhood Structure}
A number of methods have been used to further improve the performance of neural networks on subsequent problem-specific tasks, such as expressing high-dimensional data in low-dimensional space, in which capturing the structure of a given dataset may be purposely learned. This structure contains the class information of a given dataset which indicates how the class-similar data points cluster together in a projected space. Several techniques have been introduced and are popularly used for this purpose such as principal components analysis (PCA), t-stochastic neighbor embedding (t-SNE) \cite{maaten2008visualizing}, triplet loss \cite{chechik2010large}, and the soft nearest neighbor loss \cite{agarap2020improving, frosst2019analyzing, salakhutdinov2007learning} to transform the input features to contain attributes that are primed for classification.

The aforementioned techniques capture the underlying class neighborhood structure of an input data. However, each technique has their respective drawbacks. If the most salient features of the data can be found in a nonlinear manifold, then a linear technique like PCA will not be able to fully learn the underlying structure of the data. On the other hand, nonlinear techniques like t-SNE, soft nearest neighbor loss, and triplet loss require relatively expensive computational cost. Not to mention that t-SNE discovers different representations of an input data as a function of its hyperparameters. Meanwhile, the soft nearest neighbor loss and triplet loss can be relatively much slower to compute even when compared to t-SNE, depending on the available compute resources.

Despite the known drawbacks, it is still desirable to learn the underlying structure of a data as it implies how the input features form clusters, and consequently, these clusters imply the class membership of the input features as per the clustering assumption in the semi-supervised learning literature \cite{chapelle2009semi}. Although it is established that nonlinear techniques are relatively slower and more computationally expensive to use, the evolution of computational hardware has provided drastic advancements that compute issues might be negligible for those with the necessary resources.

\subsection{Ensemble Learning}

Aside from learning the class neighborhood structure of a given data, we can improve the performance of neural networks by combining their outputs through averaging or summation \cite{breiman1996stacked}. This technique is known as ensemble learning, which enforces \textit{cooperation} among neural networks to solve a common goal. In this context, we define \textit{cooperation} as the phenomenon where the neural networks in an ensemble contribute to the overall success of the group. The neural networks in an ensemble cooperate among themselves by compensating for the performance of one another, thereby decreasing the error correlation of the group.

In contrast, if a training data can be naturally divided into subsets, the group of neural networks can be inspired to specialize on their own subsets rather than to cooperate among themselves. This approach is known as the mixture-of-experts model \cite{jacobs1991adaptive} which uses a gating network as a supervisor to choose which sub-network must be assigned on a subset of the dataset based on their predictive performance on such subsets.

However, if the raw input features fed to the gating network contain highly entangled class boundaries, the gating network struggles to confidently partition the space. This often leads to ``expert collapse'' where a single expert dominates the task or multiple experts learn redundant representations. In such a case, the gating network often results to hard-routing strategies that fail to generalize well on complex datasets.

\subsection{Contributions}
To address the issue of expert collapse, we propose an architectural pipeline that primes the input features to the MoE model using SNNL. Our core contributions are:

\begin{enumerate}
    \item \textbf{Architectural Enhancement}. We employ a feature extractor regularized with SNNL before the MoE routing phase, thus minimizing intra-class distances to simplify the gating network's partition task.
    \item \textbf{Quantitative Specialization Metrics}. We address the need for detailed analysis by introducing Expert Specialization Entropy and Pairwise Embedding Similarity to empirically measure expert divergence and routing flexibility.
    \item \textbf{Statistical Rigor and Visual Proof}. We conduct robust non-parametric statistical testing (Wilcoxon signed-rank test) and provide comprehensive manifold and embedding visualizations to definitively prove that SNNL prevents structural collapse, thus allowing experts to be utilized more collaboratively on complex benchmark datasets.    
\end{enumerate}

\section{Background}

\subsection{Feed-Forward Neural Network}
The feed-forward neural network is the quintessential deep learning model that is used to approximate the  function mapping between the dataset input and output, i.e. $y \approx f\left( \Vec{x}; \theta \right)$. Its $\theta$ parameters are then optimized to learn the best approximation for the input targets, which may either be a class label (classification) or a real value (regression).
\begin{align}\label{eq:hidden_layers}
    f(\Vec{x}) = f^{(n)}\left( f^{(\ldots)} f^{(1)}(\Vec{x}) \right)
\end{align}
\begin{align}\label{eq:cross_entropy}
    \ell_{\text{ce}}(y, f(x)) = -\sum_{i} y_{i} \log \left[ f(x_{i}) \right]
\end{align}
\indent To accomplish this, the model composes a number of nonlinear functions in the form of hidden layers, each of which learns a representation of the input features (see Eq. \ref{eq:hidden_layers}). Afterwards, the similarities between the approximation and the input targets are measured by using an error function such as cross entropy (see Eq. \ref{eq:cross_entropy}), which shall be the basis for optimization usually through gradient-based learning.

For our experiments, we use a feed-forward neural network with a single hidden layer containing 128 units. The hidden layer weights were initialized with Kaiming initializer \cite{he2015delving} and had ReLU \cite{nair2010rectified} activation function while the output layer was initialized with Xavier initializer \cite{glorot2010understanding}. We use this architecture for the expert networks in our MoE models. For the gating network, we simply use a linear classifier to select which expert to use for particular input features.

\subsection{Convolutional Neural Network}
The convolutional neural network (or CNN) is a neural network architecture that uses the convolution operator as its feature extractor in its hidden layers \cite{fukushima1979neural, lecun1998gradient}. Like feed-forward neural networks, they also compose hidden layer representations for a downstream task. However, with its use of the convolution operator in its hidden layers, it learns better representations of an input data.

\begin{figure}[h]
    \centering
    \includegraphics[width=\linewidth]{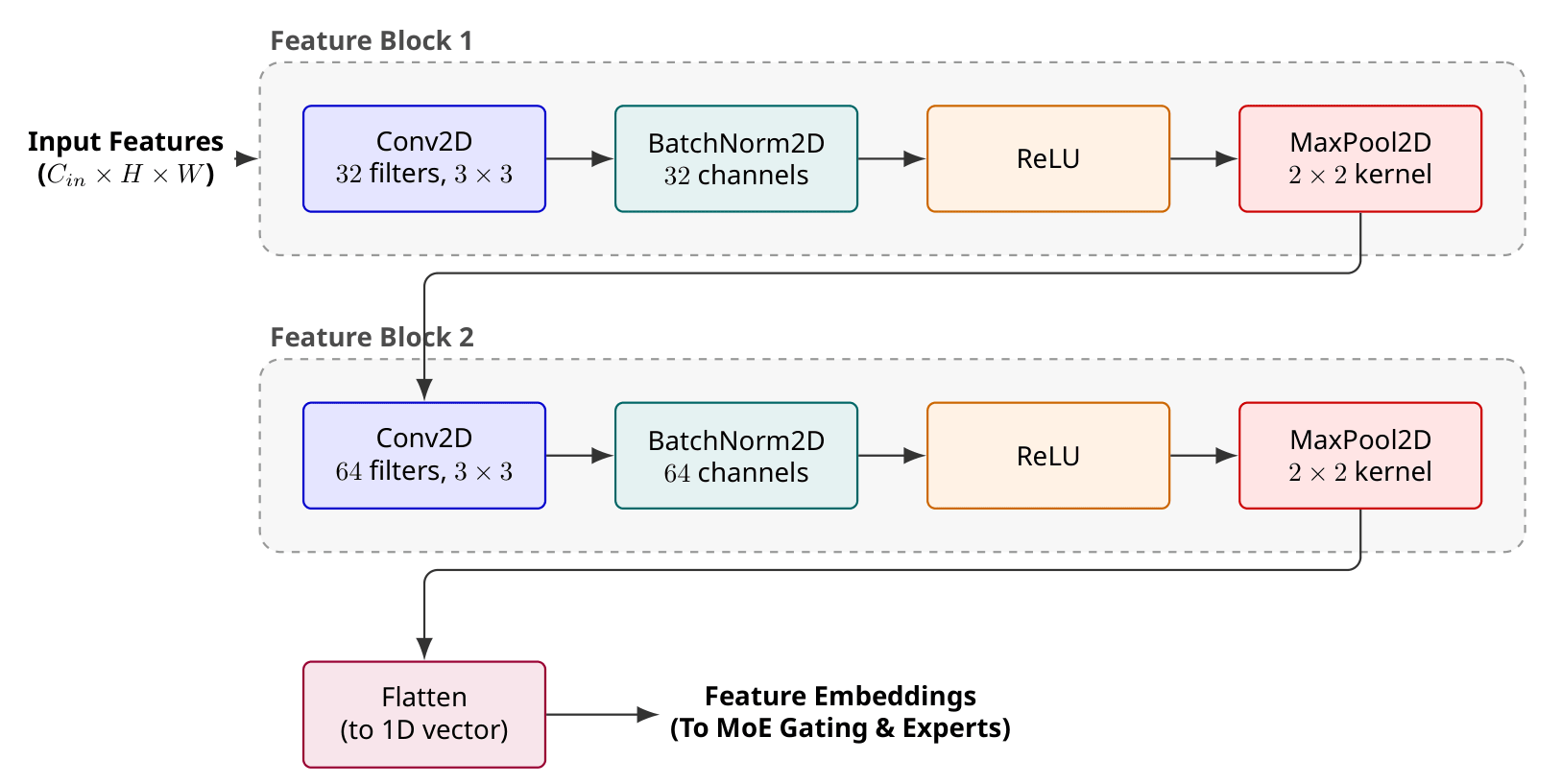}
    \caption{Architecture of the CNN-based Feature Extractor. This module computes disentangled representations that are subsequently routed to the MoE gating network and experts.}
    \label{fig:cnn-feature-extractor}
\end{figure}

In Figure \ref{fig:cnn-feature-extractor}, we illustrate the CNN-based feature extractor we use to prime the input features for the gating and expert networks across all our experiments. The network processes raw input features through a sequence of two distinct feature blocks to construct a disentangled latent representation. Each block consists of a 2D convolutional layer utilizing a $3 \times 3$ kernel with 32 filters in the first block and 64 filters in the second, then immediately followed by 2D batch normalization, a ReLU activation function to introduce non-linearity, and a $2 \times 2$ max pooling layer to progressively downsample the spatial dimensions. After the second block, the resulting multi-channel feature maps are flattened into a concise 1D vector. This final embedding serves as the topologically structured, SNNL-optimized input fed directly into the downstream MoE gating network and expert modules.

\subsection{Mixture-of-Experts}\label{sec:moe}

\begin{figure}[htb!]
    \centering
    \includegraphics[width=\linewidth]{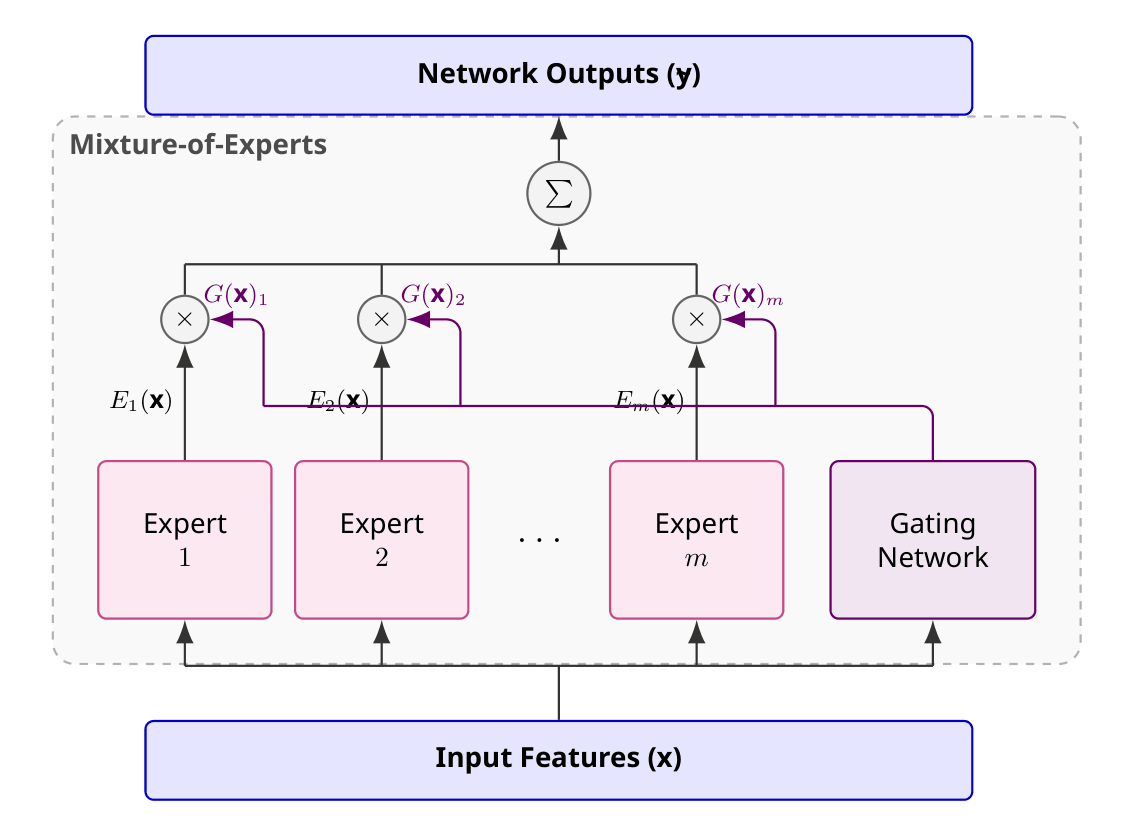}
    \caption{The mixture-of-experts model is a system of experts and gating networks where each expert become a function of a subset of the input environment. These expert networks receive the same inputs and produce the same number of outputs. The gating network also receives the same input as the expert networks, but its output is the probability of choosing a particular expert on a given input.}
    \label{fig:mixture_of_experts}
\end{figure}

The Mixture-of-Experts (MoE) model consists of a set of $n$ ``expert'' neural networks $E_{1}, \ldots, E_{n}$ and a ``gating'' neural network $G$. Figure \ref{fig:mixture_of_experts} is an illustration of the MoE model. The experts are  chosen by the gating network to handle a subset of the entire dataset, wherein each subset is tantamount to a sub-region of the data space. The output of this model is given by the following equation,
\begin{align}
    \hat{y} = \sum_{i = 1}^{n} G(x)_{i}E_{i}(x)
\end{align}
where $G(x)_{i}$ is the probability output of the gating network to choose expert $E_{i}$ for a given input $x$. The gating network and the experts have their different set of parameters $\theta$. In our classification experiments, we modified the inference function above as follows,
\begin{align}
    \hat{y} = \sum_{i = 1}^{n} \arg\max G(x)_{i} E_{i}(x)
\end{align}
This modified inference function allows the selection of the best expert network $E_{i}$ as indicated by the highest gating network output $\arg\max G(x)_{i}$. Without this modification, the model only results to a model akin to a traditional ensemble model albeit outputs a weighted summation as its output instead of a simple summation of expert outputs.\\
\indent Subsequently, we optimize the MoE model based on the following error function,
\begin{align}
    \mathcal{L}_{\text{moe}}(x, y) = \dfrac{1}{b} \sum_{i = 1}^{n} G(x)_{i} \times \ell_{ce}(y, E_{i}(x))
\end{align}
where $\ell_{\text{ce}}$ is the cross entropy function measuring the difference between the target $y$ and the output of expert $E_i$, while $G(x)_{i}$ is the probability output for choosing expert $i$. Then, the loss is averaged over the number of batch samples $b$.

In this system, each expert learns to specialize on the cases where they perform well, and they are imposed to ignore the cases on which they do not perform well. With this learning paradigm, the experts become a function of a subregion of the data space, and thus their set of learned weights highly differ from each other as opposed to traditional ensemble models that result to having almost identical weights for their learners.

\section{Resolving Expert Collapse with Disentanglement}

\subsection{Soft Nearest Neighbor Loss}\label{sec:snnl}
We define \textit{disentanglement} as how close pairs of class-similar data points from each other are, relative to pairs of class-different data points, and we can measure this by using the soft nearest neighbor loss (SNNL) function \cite{frosst2019analyzing, salakhutdinov2007learning}.

This loss function is an expansion on the original nonlinear neighborhood components analysis objective which minimizes the distances among class-similar data points in the latent code of an autoencoder network \cite{salakhutdinov2007learning}. On the other hand, the SNNL function minimizes the distances among class-similar data points in each hidden layer of a neural network \cite{frosst2019analyzing}. 
The SNNL function is defined for a batch of $b$ samples $(x, y)$ as follows,
\begin{equation}\label{eq:snnl}
    \ell_{\text{snn}}(x, y, T) = -\dfrac{1}{b} \sum_{i \in 1 \dots b} \log \left( \dfrac{\sum\limits_{\substack{j \in 1 \dots b \\ j \neq i \\ y_{i} = y_{j}}} \exp\left(-\dfrac{d_{ij}}{T}\right) }{\sum\limits_{\substack{k \in 1 \dots b \\ k \neq i}} \exp\left(-\dfrac{d_{ik}}{T}\right)} \right)
\end{equation}
where $d$ is a distance metric on either raw input features or learned hidden layer representations $x$ of a neural network, and $T$ is a temperature parameter that can be used to influence the value of the loss function. That is, at high temperatures, the distances among widely separated data points can influence the loss value.

\subsection{Our approach}
\begin{figure}
    \centering
    \includegraphics[width=\linewidth]{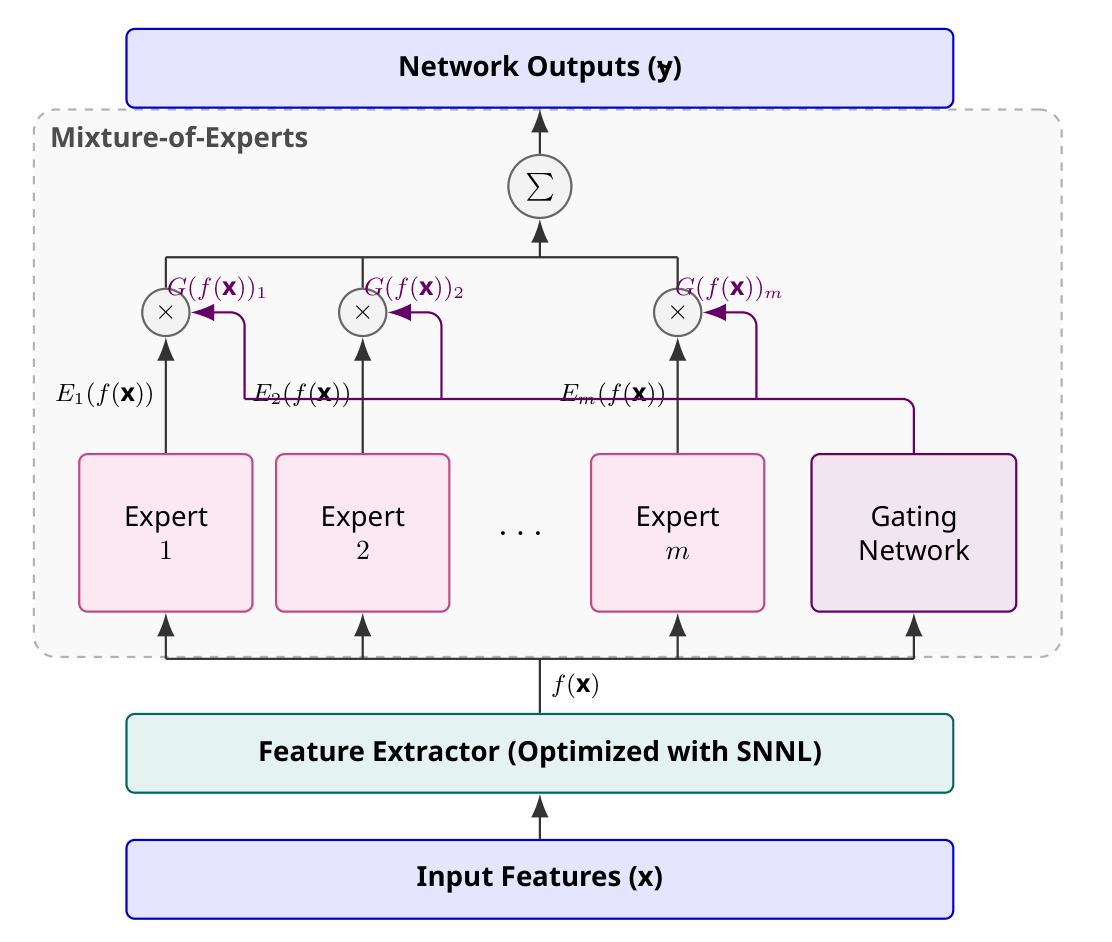}
    \caption{We optimize the soft nearest neighbor loss over the hidden layers found in the feature extractor network before the MoE model. In doing so, the input features to the expert and gating networks are transformed to a set of representations with the classification information ingrained in them, thereby helping improve the overall classification performance of the MoE model.}
    \label{fig:snnl_moe}
\end{figure}

We propose to use a feature extractor network $f$ for the MoE model instead of using the raw input features for the expert and gating networks (see Figure \ref{fig:snnl_moe}). Using our feature extractor network optimized with SNNL, we prime the input features via disentanglement w.r.t. their class labels. Our proposal thus leads to the optimization of a composite loss (see Eq. \ref{eq:composite}) of the cross entropy as the classification loss and the SNNL as the regularizer.
\begin{align}\label{eq:composite}
    \mathcal{L}(f, x, y) = \mathcal{L}_{\text{moe}} (x, y) + \alpha \cdot \min_{\ell_{\text{snn}}} \ell_{\text{snn}} (f^{i}(x), y, T)
\end{align}
We found more stable and better optimization of the SNNL when we take its minimum value across the hidden layers $f^{i}$ of the feature extractor network.

\subsection{Theoretical Analysis}
To formalize the contribution of the soft nearest neighbor loss (SNNL) to expert specialization, we analyze the interaction between the feature extractor $f$ and the Mixture-of-Experts (MoE) gating mechanism. 

\subsubsection{Latent Space Partitioning via SNNL}
The SNNL objective $l_{snn}$ serves as a topological regularizer on the feature extractor's hidden layers. By minimizing the distance between class-similar data points relative to class-different ones, the feature extractor $f(\vec{x})$ maps the high-dimensional input $\vec{x}$ into a lower-dimensional latent manifold where class clusters are highly localized. Mathematically, the SNNL gradient w.r.t. the weights of $f$ minimizes the pairwise distance $d_{ij}$ for all $y_i = y_j$. This results in a latent representation $z = f(\vec{x})$ where the intra-class variance is minimized.

\subsubsection{Gating Dynamics and Expert Selection}
In a traditional MoE model, the gating network $G(\vec{x})$ often suffers from ``expert collapse'' where a single expert dominates the gradients due to a disorganized input space. In our approach, the gating network operates on the transformed representation $G(f(\vec{x}))$. 

Since $f(\vec{x})$ essentially clusters inputs by class, the gating function $G(z)$ encounters a ``pre-partitioned'' environment. Let $z_c$ be the centroid of class $c$ in the latent space. The gating network learns a mapping $\phi : z_c \to \{1, \dots, n\}$, effectively assigning a specific expert $E_i$ to a specific class cluster. As the inference function utilizes $\arg \max G(z)_i$, the model selects the "best" expert for a given sub-region of the data space.

\subsubsection{Formalizing the Specialization}
Specialization is achieved when the variance of the input distribution seen by expert $E_i$ is significantly lower than the global distribution variance:
$$\text{Var}(f(\vec{x}) | G(f(\vec{x}))_i \approx 1) \ll \text{Var}(f(\vec{x}))$$

By ensuring that the input to the gating network is already disentangled by class, our approach forces each expert $E_i$ to learn a localized function $f^{(E_i)}$ specifically tuned to the nuances of the assigned class cluster. This reduces the interference between experts and facilitates higher classification accuracy, as evidenced by the empirical results on the classification datasets used in the study.


\section{Experiments}

We use four benchmark image classification datasets to evaluate our proposed model: MNIST, FashionMNIST, CIFAR10, and CIFAR100. We ran each model five times, and computed their average performance across those runs. We report both the average classification performance and the best classification performance for each of our model. For reproducibility, we used the following set of seeds for the random number generator: 1, 2, 3, 4, and 5. No hyperparameter tuning was done as trying to achieve state-of-the-art performance is beyond the scope of this study, we only intend to show that using a feature extractor optimized with SNNL helps resolve the issue of expert collapse in a mixture-of-experts model. In addition, no other regularizers were used in order to better demonstrate the benefits of SNNL for our feature extractor.

\subsection{Evaluation Metrics}
To thoroughly evaluate the impact of SNNL and address the limitations of purely accuracy-based metrics, we employed the following analytical framework:

\begin{enumerate}
    \item \textbf{Expert Specialization Entropy (ENT)}. We calculate the Shannon entropy of the average routing distribution per class. Lower entropy indicates ``hard-routing'' while higher entropy indicates a more flexible routing distribution across the experts.
    \item \textbf{Pairwise Embedding Similarity (SIM)}. We extract the first-layer weight matrices of each expert network and compute their pairwise cosine similarity. Lower similarity indicates that the experts have learned orthogonal and divergent internal representations, thus escaping expert collapse.
    \item \textbf{Statistical Analysis}. To verify if the improvements were robust, all models were trained across multiple random seeds. The differences in metrics were evaluated using the non-parametric Wilcoxon signed-rank test.
    \item \textbf{Manifold Visualization}. We project the high-dimensional feature representations in the feature extractor into 2D space using UMAP \cite{mcinnes2018umap} to qualitatively assess cluster homogeneity and confirm the prevention of structural collapse.
\end{enumerate}

\subsection{Experimental Setup}

To ensure our findings are applicable in resource-constrained research environments, all models were trained locally using PyTorch Lightning on an RTX 3060 GPU. Since the SNNL computation is only required during the training phase to shape the feature extractor's weights, our proposed architectural enhancement adds zero computational overhead during inference.

We trained all our models for 15,000 steps on a mini-batch size of 100 using SGD with momentum (0.9) and weight decay (1e-4) \cite{polyak1964some} with a learning rate of 1e-1, and we used OneCycleLR to anneal the learning rate \cite{smith2017super}.

\subsection{Classification Performance}
Table \ref{tab:classification-results} summarizes the performance across all three core metrics. The experimental models demonstrate significant behavioral shifts compared to the baselines, i.e. trading rigid routing for structural diversity which particularly benefited the more complex datasets.

\begin{table*}[]
\centering
\resizebox{\textwidth}{!}{%
\begin{tabular}{|c|cc|cc|cc|}
\hline
\rowcolor[HTML]{F2F2F2} 
\cellcolor[HTML]{F2F2F2}                                   & \multicolumn{2}{c|}{\cellcolor[HTML]{F2F2F2}\textbf{ACC (MEAN (SD) in \%)}}                 & \multicolumn{2}{c|}{\cellcolor[HTML]{F2F2F2}\textbf{SIM (MEAN (SD) in x100)}}          & \multicolumn{2}{c|}{\cellcolor[HTML]{F2F2F2}\textbf{ENT (MEAN (SD) in x100)}}           \\ \cline{2-7} 
\rowcolor[HTML]{F2F2F2} 
\multirow{-2}{*}{\cellcolor[HTML]{F2F2F2}\textbf{Dataset}} & \multicolumn{1}{c|}{\cellcolor[HTML]{F2F2F2}\textbf{Baseline}} & \textbf{Experimental}      & \multicolumn{1}{c|}{\cellcolor[HTML]{F2F2F2}\textbf{Baseline}} & \textbf{Experimental} & \multicolumn{1}{c|}{\cellcolor[HTML]{F2F2F2}\textbf{Baseline}} & \textbf{Experimental}  \\ \hline
MNIST                                                      & \multicolumn{1}{c|}{\textbf{99.36 (0.03) *}}                   & 99.25 (0.08)               & \multicolumn{1}{c|}{\textbf{0.03 (0.03)}}                      & 0.04 (0.06)           & \multicolumn{1}{c|}{\textbf{0.58 (0.06)}}                      & 1.94 (0.14)            \\ \hline
FMNIST                                                     & \multicolumn{1}{c|}{91.33 (0.13)}                              & \textbf{91.61 (0.23) *}    & \multicolumn{1}{c|}{0.03 (0.07)}                               & \textbf{-0.01 (0.05)} & \multicolumn{1}{c|}{16.82 (10.88)}                             & \textbf{16.00 (11.03)} \\ \hline
CIFAR10                                                    & \multicolumn{1}{c|}{70.91 (1.09)}                              & \textbf{71.23 (0.91) (ns)} & \multicolumn{1}{c|}{0.05 (0.11)}                               & \textbf{0.03 (0.06)}  & \multicolumn{1}{c|}{\textbf{14.00 (15.28)}}                    & 15.01 (15.52)          \\ \hline
CIFAR100                                                   & \multicolumn{1}{c|}{35.75 (1.64)}                              & \textbf{36.74 (1.59) *}    & \multicolumn{1}{c|}{0.20 (0.24)}                               & \textbf{0.10 (0.15)}  & \multicolumn{1}{c|}{\textbf{43.92 (26.88)}}                    & 45.74 (26.60)          \\ \hline
\end{tabular}%
}
\caption{Classification Accuracy (ACC), Embedding Similarity (SIM), and Routing Entropy (ENT) across Baseline and Experimental MoE models. Values represent MEAN (SD). Asterisks (*) denote statistical significance at $p < 0.05$ via Wilcoxon signed-rank test; (ns) denotes not significant.}
\label{tab:classification-results}
\end{table*}

\textbf{The Complex Datasets (FashionMNIST and CIFAR100)}. The experimental model shows a statistically significant improvement over the Baseline in classification accuracy. This confirms that for more difficult and highly entangled feature spaces, priming the latent space with SNNL genuinely aids downstream classification.

\textbf{The Simple Dataset (MNIST)}. The baseline model significantly outperforms the experimental model. Since MNIST is an extremely simple and highly separable dataset (baseline accuracy approaches 99.4\%), we suspect forcing SNNL clustering likely introduces over-regularization or unnecessary constraints that marginally degrade performance.

\subsection{Ablation and Specialization Analysis}

Our specialization metrics potentially reveals why the performance on complex datasets improved with our experimental approach. The quantitative data contradicts the assumption that SNNL merely enforces harder routing; instead, SNNL fundamentally prevents structural expert collapse.

\textbf{Orthogonal Embeddings (Resolving Expert Collapse)}. On FashionMNIST, CIFAR10, and CIFAR100, the SNNL model achieved noticeably lower embedding similarities. This is profoundly pronounced on CIFAR100, where the baseline experts suffered from high similarity (0.20), indicating structural redundancy and collapse. SNNL halved this redundancy (0.10). On FashionMNIST, it pushed the similarity into the negative (-0.01).

\begin{figure}[h]
    \centering
    \includegraphics[width=\linewidth]{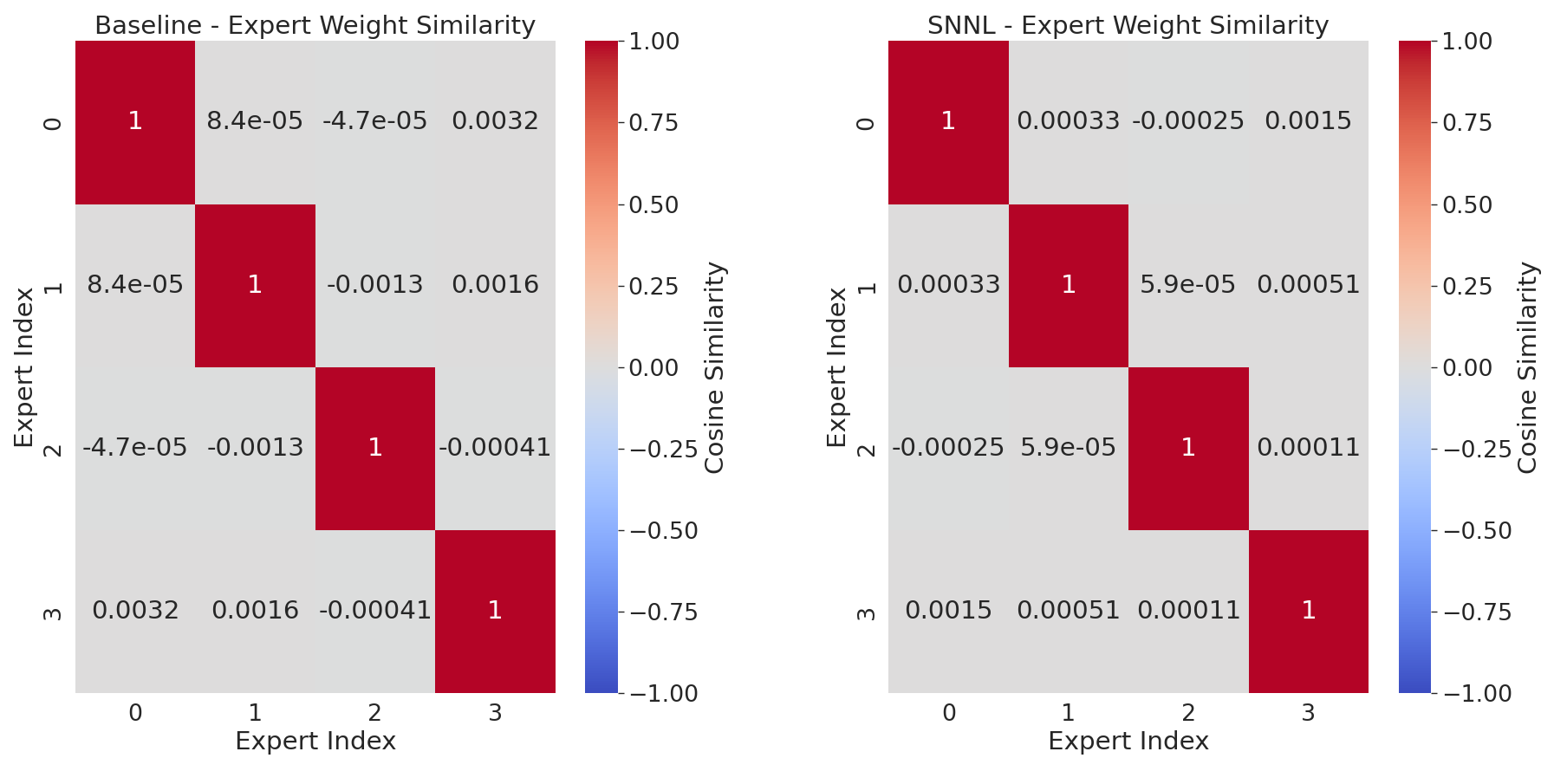}
    \caption{Pairwise cosine similarity between expert weight matrices on CIFAR10. (Left) The Baseline shows distinct patches of redundancy (e.g. between Expert 0 and 3). (Right) The SNNL condition visibly suppresses these redundancies, thus forcing experts to learn highly distinct and orthogonal features.}
    \label{fig:cifar10_expert_similarity}
\end{figure}

As visualized in Figure \ref{fig:cifar10_expert_similarity}, the baseline condition exhibits redundancy among experts. By feeding disentangled representations to the MoE, the experts are forced to learn highly distinct and non-overlapping weights, thus cutting the highest off-diagonal similarity by more than half.

\textbf{Routing Entropy (Flexible Collaboration)}. Our results show that aside from FashionMNIST, the SNNL model actually increased routing entropy. The SNNL feature extractor makes the latent clusters so well-defined that the gating network does not need to hard route input features. Instead, it employs a more distributed routing strategy (higher entropy) while relying on the structurally diverse, orthogonal experts to handle classification nuances.

\begin{figure}
    \centering
    \begin{subfigure}[h]{0.99\linewidth}
        \includegraphics[width=\linewidth]{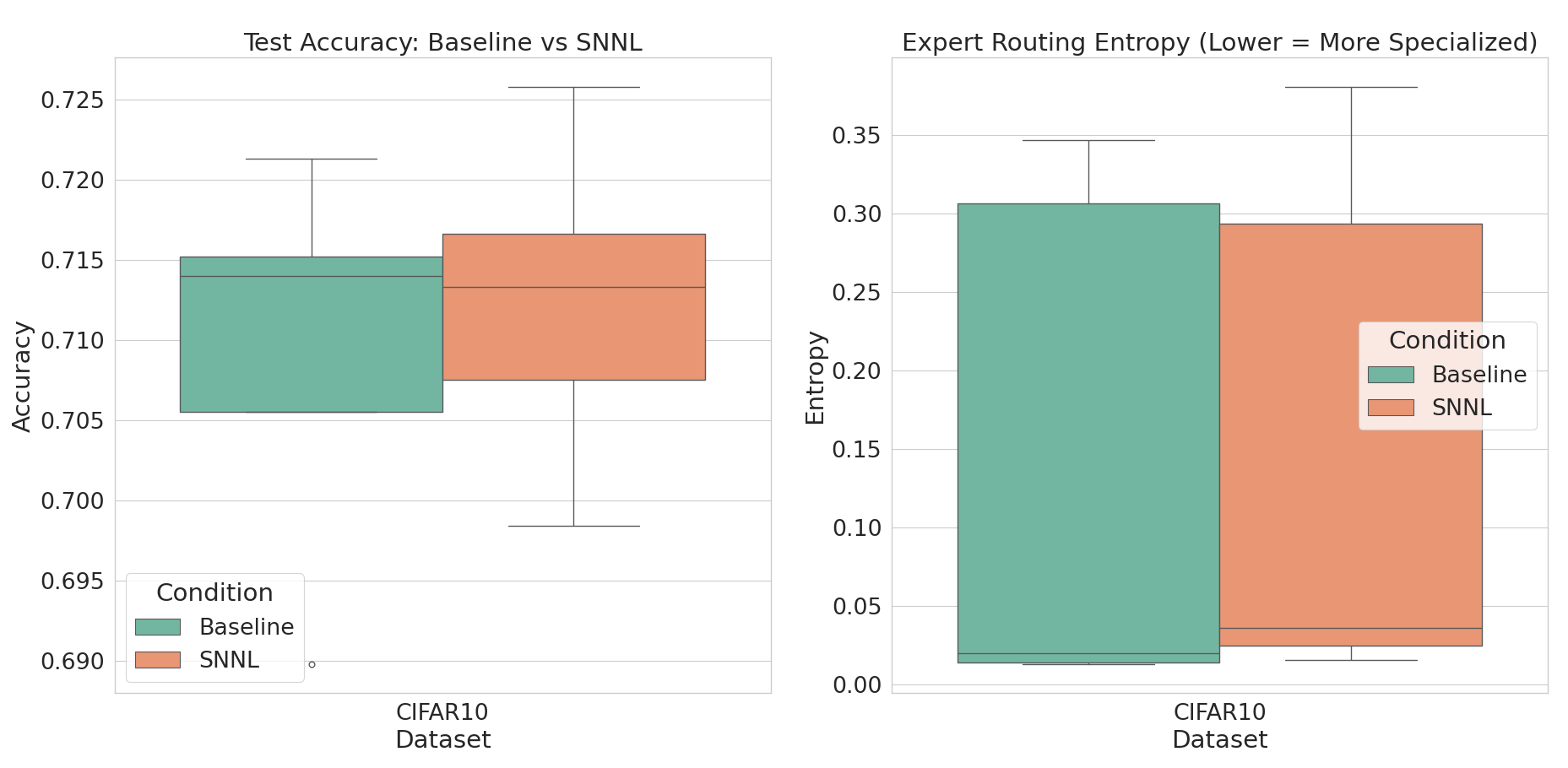}
    \end{subfigure}
    \begin{subfigure}[h]{0.99\linewidth}
        \includegraphics[width=\linewidth]{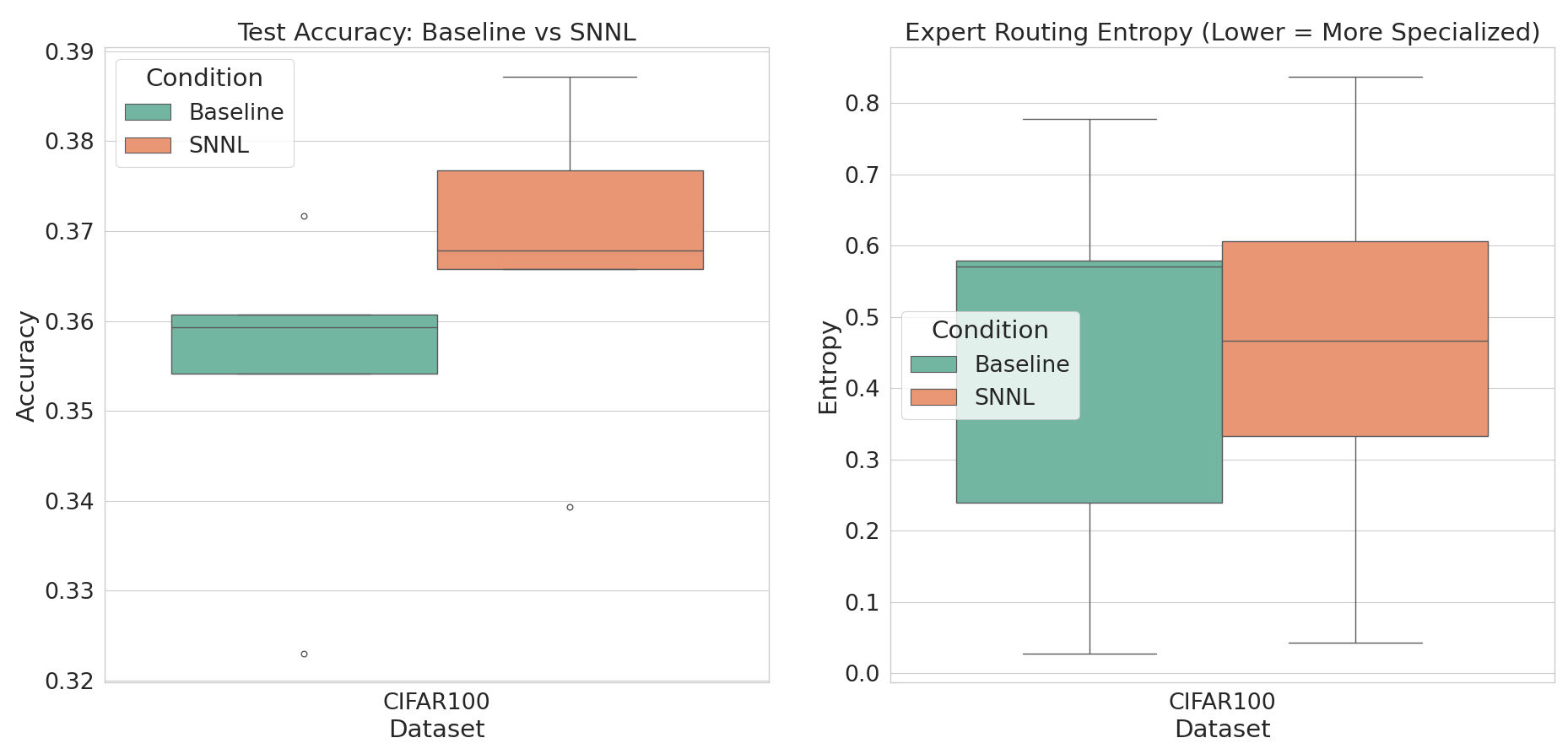}
    \end{subfigure}
    \caption{Distribution of Test Accuracy and Routing Entropy across random seeds. (Top/CIFAR100) SNNL provides a robust, highly stable boost to both accuracy and entropy, elevating the entire interquartile range. (Bottom/CIFAR10) While median entropy slightly increases, the wider variance and overlapping accuracy boxes explain the non-significant accuracy gain on this intermediate dataset.}
    \label{fig:boxplots}
\end{figure}

Figure \ref{fig:boxplots} visualizes our ``flexible collaboration'' narrative. On the highly complex CIFAR100 dataset, the SNNL condition elevates the entire interquartile range for accuracy, proving a robust performance boost while simultaneously raising the median entropy. On the intermediate CIFAR10 dataset, the SNNL boxplot reaches a higher maximum accuracy, but the heavily overlapping boxes contextualize why the overall gain was not statistically significant. Nevertheless, the underlying behavioral shift towards higher entropy remains intact across both complexity levels.

\begin{figure}[h]
    \centering
    \includegraphics[width=\linewidth]{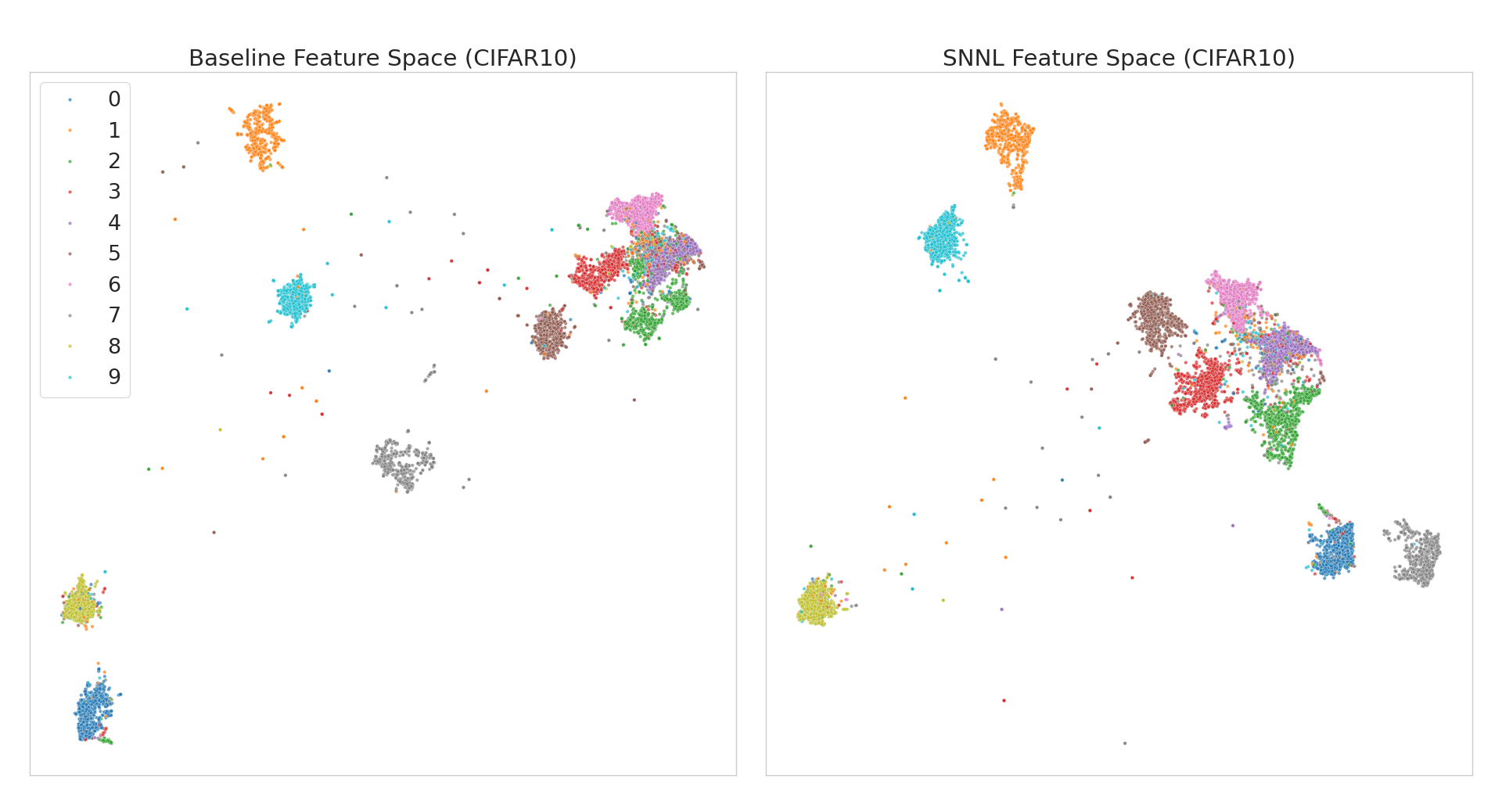}
    \caption{UMAP projection of the latent feature space immediately prior to the MoE gating network on CIFAR10. (Left) The baseline feature extractor produces highly entangled class boundaries. (Right) The SNNL-regularized feature extractor minimizes intra-class distance, producing dense and homogeneous class clusters.}
    \label{fig:cifar10_umap}
\end{figure}

\textbf{Manifold Visualization (Validating Disentanglement).} The UMAP projections in Figure \ref{fig:cifar10_umap} qualitatively confirm the mechanism driving our quantitative metrics. The baseline feature space (Left) exhibits highly entangled class boundaries, which explains why baseline MoE models suffer from higher embedding similarity: the gating network is forced to send overlapping distributions of data to multiple experts. Conversely, the SNNL feature space (Right) demonstrates that intra-class variance has effectively collapsed, resulting in dense and distinct clusters. This proves that SNNL effectively disentangles the feature representations, thus trivializing the gating network's routing task and enforcing strict structural specialization.

\subsection{SNNL Weight Ablation: Tuning Expert Diversity}
To further understand the continuous effect of the SNNL weighting factor ($\alpha$) on expert orthogonality, we conducted an ablation study using a 25k subset of the CIFAR10 training set. We evaluated 30 models with $\alpha$ values linearly spaced from -100 to 100.
\begin{figure}[h]
    \centering
    \includegraphics[width=\linewidth]{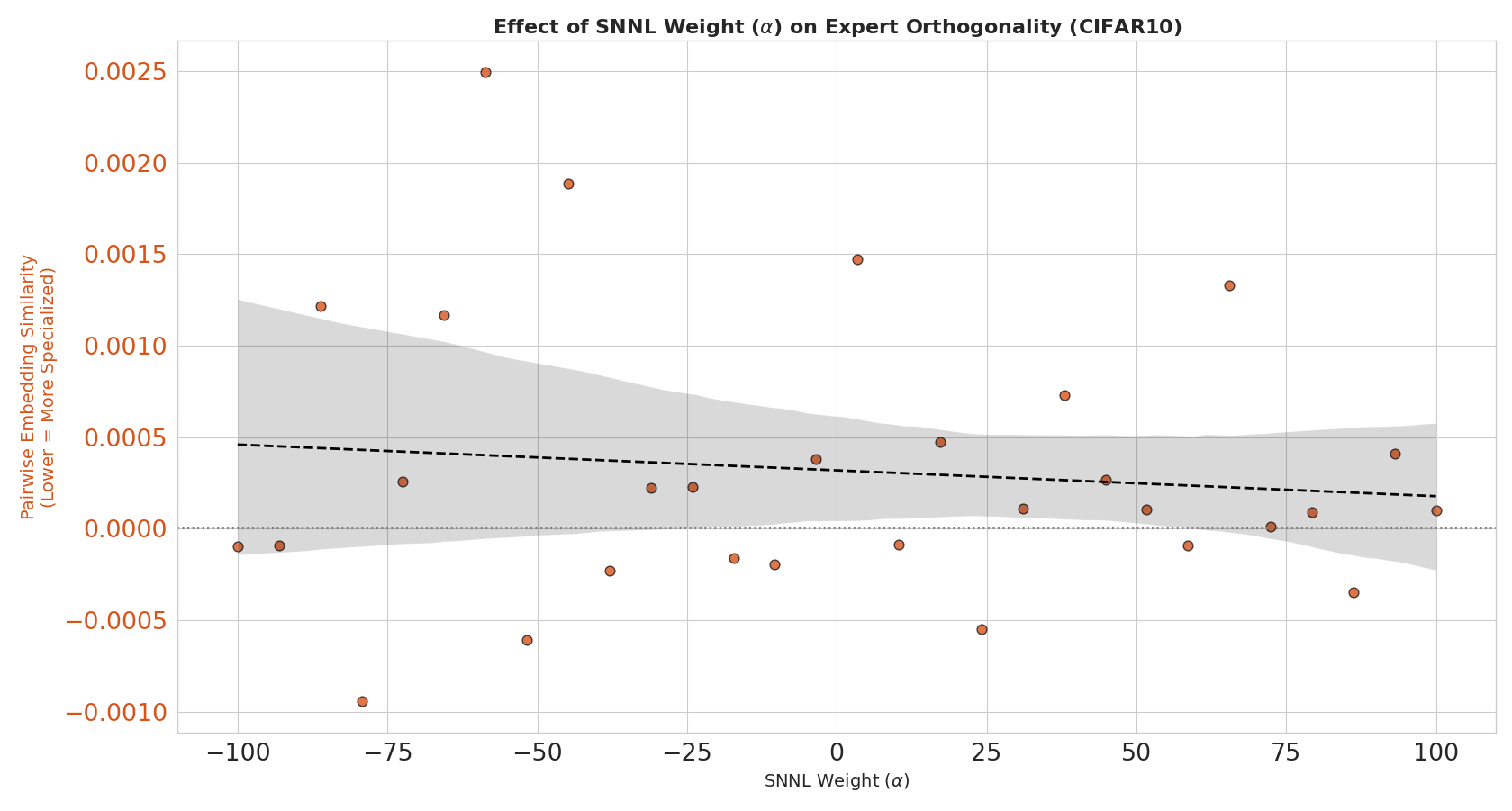}
    \caption{The effect of the SNNL weight ($\alpha$) on the pairwise embedding similarity of the experts. The dashed regression line shows a negative correlation, indicating that stronger SNNL penalties actively suppress expert redundancy.}
    \label{fig:snnl_weight_specialization}
\end{figure}

As illustrated in Figure \ref{fig:snnl_weight_specialization}, the trend line confirms a negative correlation between the SNNL weight and pairwise embedding similarity among the experts. While MoE routing exhibits inherent variance across runs, progressively increasing $\alpha$ actively pulls the expected similarity closer to 0. This demonstrates that the SNNL weight acts as a tunable control parameter, i.e. a stronger topological penalty directly forces the downstream experts to learn more distinct, non-overlapping representations.

\section{Recommendations}
While our findings demonstrate the significant advantages of SNNL in resolving expert collapse, several avenues for future research remain to fully realize the potential of topology-aware MoE architectures:

\begin{enumerate}
    \item \textbf{Dynamic Parameter Tuning}. Our ablation study highlighted the sensitivity of expert orthogonality to the SNNL weighting factor ($\alpha$). Future work should explore dynamically adjusting both $\alpha$ and the temperature parameter ($T$) during training. Utilizing learning rate schedulers to gradually decay or warm up these values could better optimize the transition from global representation clustering to fine-grained cross-entropy classification.
    \item \textbf{Architectural Scaling}. The current experiments were constrained to lightweight CNN variants to accommodate local computing limits. Future investigations should scale this methodology to Vision Transformers (ViTs) and higher-resolution, large-scale datasets (e.g. ImageNet) to verify if the disentanglement benefits scale linearly with model size and extreme class diversity.
    \item \textbf{Comparative Representation Learning}. Direct benchmarking against other modern contrastive learning techniques such as Supervised Contrastive Learning (SupCon) \cite{khosla2020supervised} or SimCLR \cite{chen2020simple} is necessary. This will isolate whether SNNL's specific layer-wise distance minimization is uniquely suited for MoE gating or if any class neighborhood structure-preserving clustering objective yields similar structural diversity and routing flexibility.

\end{enumerate}

\newpage
\section{Conclusion}
We proposed and evaluated an architectural modification to the Mixture-of-Experts (MoE) model by integrating a feature extractor optimized via Soft Nearest Neighbor Loss (SNNL). By expanding our evaluation to include specialization metrics like Pairwise Embedding Similarity and Expert Specialization Entropy, and including UMAP manifold visualization, we uncovered a nuanced mechanism of action that addresses the limitations of standard MoE architectures.

Our empirical findings demonstrate that baseline MoE architectures often suffer from structural expert collapse, where experts learn redundant representations which results to rigid routing by the gating network. Disentangling the feature representations with SNNL solves this structural collapse by forcing the experts to learn highly orthogonal and diverse features. Consequently, this liberates the gating network to utilize a more collaborative routing strategy.

While disentanglement may have the tendency to ``over-regularize'' models on simple datasets like MNIST, it yields statistically significant accuracy improvements on complex or highly entangled datasets such as FashionMNIST, CIFAR10, and CIFAR100, thus effectively bridging the gap between representation learning and ensemble specialization.

\bibliographystyle{ACM-Reference-Format}

\bibliographystyle{ACM-Reference-Format}

\end{document}